\documentclass[11pt,a4paper]{article}
\usepackage[utf8]{inputenc}
\usepackage[T1]{fontenc}
\usepackage{cite}
\usepackage{amsmath,amssymb,amsfonts}
\usepackage{algorithmic}
\usepackage{graphicx}
\usepackage{textcomp}
\usepackage{natbib}
\usepackage{subcaption}
\usepackage{booktabs}
\usepackage{bm}
\usepackage{geometry}
\usepackage{authblk}

\makeatletter
\AtBeginDocument{\DeclareMathVersion{bold}
\DeclareSymbolFont{NewLetters}{T1}{times}{m}{it}
\SetSymbolFont{operators}{bold}{T1}{times}{b}{n}
\SetSymbolFont{NewLetters}{bold}{T1}{times}{b}{it}
\SetMathAlphabet{\mathrm}{bold}{T1}{times}{b}{n}
\SetMathAlphabet{\mathit}{bold}{T1}{times}{b}{it}
\SetMathAlphabet{\mathbf}{bold}{T1}{times}{b}{n}
\SetMathAlphabet{\mathtt}{bold}{OT1}{pcr}{b}{n}
\SetSymbolFont{symbols}{bold}{OMS}{cmsy}{b}{n}
\renewcommand\boldmath{\@nomath\boldmath\mathversion{bold}}}
\makeatother

\title{ConceptACT: Episode-Level Concepts for Sample-Efficient Robotic Imitation Learning}

\author[1]{Jakob Karalus}
\author[2]{Friedhelm Schwenker}

\affil[1]{Institute of Artificial Intelligence, Ulm University, Ulm, Germany}
\affil[2]{Institute of Neuroinformatics, Ulm University, Ulm, Germany}

\date{}

\begin{document}

\maketitle

\begin{abstract}
Imitation learning enables robots to acquire complex manipulation skills from human demonstrations, but current methods rely solely on low-level sensorimotor data while ignoring the rich semantic knowledge humans naturally possess about tasks. We present ConceptACT, an extension of Action Chunking with Transformers that leverages episode-level semantic concept annotations during training to improve learning efficiency. Unlike language-conditioned approaches that require semantic input at deployment, ConceptACT uses human-provided concepts (object properties, spatial relationships, task constraints) exclusively during demonstration collection, adding minimal annotation burden. We integrate concepts using a modified transformer architecture in which the final encoder layer implements concept-aware cross-attention, supervised to align with human annotations. Through experiments on two robotic manipulation tasks with logical constraints, we demonstrate that ConceptACT converges faster and achieves superior sample efficiency compared to standard ACT. Crucially, we show that architectural integration through attention mechanisms significantly outperforms naive auxiliary prediction losses or language-conditioned models. These results demonstrate that properly integrated semantic supervision provides powerful inductive biases for more efficient robot learning.
\end{abstract}


\noindent\textbf{Corresponding author:} Jakob Karalus (e-mail: jakob.karalus@gmx.net)

\section{Introduction}
\label{sec:introduction}

In recent years, Imitation Learning has emerged as a successful technique for teaching robots complex skills, primarily through directly learning from human demonstrations of such skills (\cite{schaal1999imitation, argall2009survey}). The advantage of learning from demonstrations (in contrast to traditional reinforcement learning or other approaches) is that no explicit reward functions or control algorithms have to be designed. The approach of learning from (expert) trajectories makes it especially valuable for tasks where good behavior is easier to demonstrate than to formalize. However, current imitation learning methods have a fundamental limitation: they mostly rely on low-level (sensorimotor) data (joint positions, images, forces, etc.) while ignoring other knowledge that humans naturally possess when demonstrating a task (e.g., \cite{zare2024survey}). 

This is in contrast to the way humans teach complex tasks to other humans. In these settings, humans naturally employ a variety of techniques to help the learner reach their goal. The umbrella term for these techniques in psychological research is "scaffolding," a metaphor for the structural supports the learner receives during their skill-building phase. For example, techniques like providing conceptual frameworks, highlighting important features, and explaining underlying principles are natural for humans to use. But they are usually employed to accelerate the learning process (\cite{van2010scaffolding, shao2023effects}). In contrast, machine learning from demonstration (i.e., Imitation Learning) is typically a black box for the human, which learns implicit associations between high-dimensional observations and actions without access to the semantic reasoning that guides human decision-making (\cite{firoozi2025foundation}). This mismatch represents a significant missed opportunity: human demonstrators possess valuable high-level knowledge about task structure, object properties, and causal relationships that could substantially improve learning efficiency if properly leveraged.

Current state-of-the-art approaches like ACT (\cite{zhao2023act}) are limited to learning from low-level data, failing to gather the conceptual understanding that humans can readily provide alongside their demonstrations. This limitation can be especially pronounced in manipulation tasks that involve complex reasoning about object properties, spatial relationships, or task constraints, which are scenarios where semantic guidance could prove most beneficial.

We address this limitation by introducing ConceptACT, an extension of the ACT architecture that integrates episode-level semantic concepts directly into the imitation learning process. Our approach leverages a key practical insight: human demonstrators naturally observe object properties, spatial relationships, and task constraints while performing demonstrations, making episode-level concept annotation straightforward and minimally expensive (in terms of annotation cost) during the collection of the demonstration data. In contrast to methods that require semantic descriptions or language instructions at deployment time, placing ongoing burdens on end users, ConceptACT requires concept information only during the one-time demonstration phase. Once training is complete, the resulting policy operates on standard low-level inputs without needing any additional semantic information during deployment.

Our technical approach enables human demonstrators to annotate episodes with high-level concepts (such as object colors, shapes, or spatial relationships) and uses these annotations as auxiliary supervision during training. By incorporating a modified \textit{Concept Transformer} (\cite{rigotti2022concept}) architecture into the standard ACT encoder, ConceptACT learns to attend to semantically meaningful concepts while predicting action sequences, creating stronger inductive biases that improve sample efficiency and task understanding. We evaluate ConceptACT on two robotic pick-and-place tasks with high-level sorting or ordering constraints, where robots must manipulate objects of varying shapes and colors according to complex conditional rules.

The primary contributions of this work are threefold: (1) we demonstrate a systematic approach for integrating episode-level semantic concepts into transformer-based imitation learning, (2) we show that proper architectural integration of concepts through attention mechanisms provides superior learning compared to auxiliary prediction tasks, and (3) we provide empirical evidence that concept-guided learning improves sample efficiency in manipulation tasks requiring conditional reasoning about object properties.

\begin{figure*}
    \centering
    \includegraphics[width=1\linewidth]{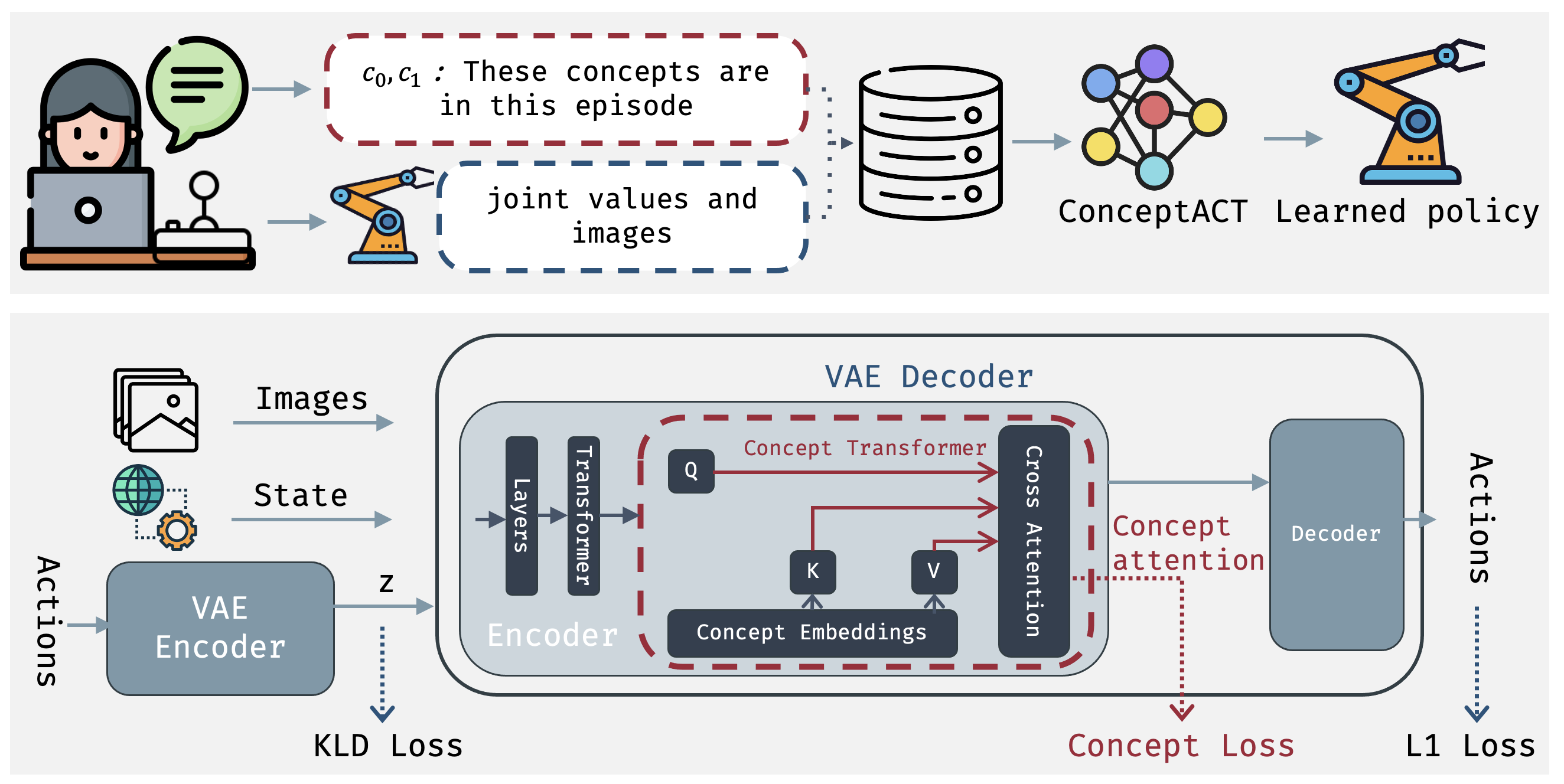}
    \caption{Overview of our approach: Top: We enhance the normal Imitation Learning approach by allowing the user to specify which concepts are in an episode. Bottom: We integrate these concepts by extending the ACT architecture to include a Concept Transformer that aligns its attention mechanism with the given concepts. This alignment cost can then be included in the total loss. Red indicates changes.}\label{fig:approach}
\end{figure*}

\section{Background}
\label{sec:background}

The problem in our sequential decision-making can be formulated as a Markov Decision Process (MDP). An MDP is defined by the tuple $(\mathcal{S}, \mathcal{A}, \mathcal{P}, \mathcal{R}, \gamma)$, where $\mathcal{S}$ represents the state space, $\mathcal{A}$ the action space, $\mathcal{P}: \mathcal{S} \times \mathcal{A} \times \mathcal{S} \rightarrow [0,1]$ the transition probability function, $\mathcal{R}: \mathcal{S} \times \mathcal{A} \rightarrow \mathbb{R}$ the reward function, and $\gamma \in [0,1]$ the discount factor.

At each timestep $t$, an agent observes state $s_t \in \mathcal{S}$, executes action $a_t \in \mathcal{A}$, and transitions to a new state $s_{t+1}$ according to the transition dynamics $\mathcal{P}(s_{t+1}|s_t, a_t)$ and receives an reward from the reward function $\mathcal{R}(s_t, a_t) -> r_t$. The agent's objective is to learn a policy $\pi: \mathcal{S} \rightarrow \mathcal{A}$ that maximizes the expected cumulative (discounted) reward. 

Since this work primarily focuses on robotics applications, the state space typically includes diverse sensory inputs, such as proprioceptive measurements and visual observations, while actions $a_t$ correspond to motor commands or target configurations. Because our work is in Imitation Learning, there is no available reward function for policy learning. 

\subsection{Imitation Learning and Behavior Cloning}

In Imitation Learning, the goal of the agents is then to learn a policy from previously collected demonstrations of the desired behaviours. This addresses scenarios where specifying an explicit reward function proves challenging or impractical, but expert demonstration of the desired behaviour can be collected. 

Given a dataset of expert demonstrations $\mathcal{D} = \{(s_i, a_i)\}_{i=1}^N$ consisting of state-action pairs collected from expert trajectories, the goal is to learn a policy $\pi_\theta$ parameterized by $\theta$ that mimics the expert's behavior. The underlying assumption is that the expert demonstrations are generated by an optimal or near-optimal policy $\pi^*$ and therefore the agent policy should mimic this behaviour as closely as possible.

To solve the problem of learning the policy $\pi$, \textit{Behavior Cloning} (BC) represents the most direct approach to the imitation learning problem, in which the problem is formulated as a simple supervised learning problem (instead of a reinforcement learning problem). The policy parameters $\theta$ are optimized to minimize the discrepancy between predicted and demonstrated actions:

\begin{equation}
\mathcal{L}_{BC}(\theta) = \mathbb{E}_{(s,a) \sim \mathcal{D}}[\ell(\pi_\theta(s), a)]
\end{equation}

where $\ell(\cdot, \cdot)$ is an appropriate loss function, typically $L_1$ or $L_2$ norm for continuous action spaces, or cross-entropy for discrete actions.

\subsection{Action Chunking with Transformers (ACT)}
%TODO: Add imagenet encoder somewhere (with citation)

One of the more prominent recent advances in Imitation learning Learning, especially for robotic manipulation, is \textit{ACT}: \textit{Action Chunking with Transformers} \citep{zhao2023act}. ACT introduces three key architectural innovations that improve policy learning from human demonstrations.
First, rather than predicting single actions at each timestep, ACT predicts sequences of actions (chunks). Second, it employs a variational autoencoder (VAE) training framework to encode style differences. Third, the integration of a transformer encoder-decoder architecture. This approach has proven particularly effective for fine-movement tasks that require temporal consistency and precise control.

\subsubsection{Action Chunking}

The core function of the policy in traditional behavior cloning (and also in reinforcement learning) is to let the policy predict the next actions, given the current state:  $\pi_\theta(a_t|s_t)$. With the introduction of the Action Chunking in ACT, the policy instead learns to predict an action sequence:

\begin{equation}
\pi_\theta(a_{t:t+k-1}|s_t) = \pi_\theta(a_t, a_{t+1}, \ldots, a_{t+k-1}|s_t)
\end{equation}

The effect of action chunking is that it reduces the effective horizon of a task from $T$ timesteps to $\lceil T/k \rceil$. This reduction can mitigate error accumulation. Action chunking also helps model non-Markovian behavior in human demonstrations, such as natural pauses that single-step policies struggle to handle.

\subsubsection{Variational Training}

The second key part of ACT is the usage of a variational autoencoder (\cite{kingma2014autoencoding}) as an additional loss for the training. Therefore, the architecture is split into two parts: a VAE-Encoder that produces a latent variable and a VAE-Decoder \footnote{We keep using a VAE prefix for the Encoder and Decoder here, to minimize naming confusion for later parts of the architecture, where there is also a transformer-based encoder and decoder}. In ACT, the VAE-Encoder aims to produce a suitable latent "style" variable $z \in \mathbb{R}^L$ that captures the characteristics of the current trajectory. This VAE-Encoder is trained on the entire sequence of joint variables as input (excluding other inputs, such as images, to minimize computational needs) and produces the latent variable $z$ as the output. At training time, a KL-loss (\cite{kullback1951information}) is computed between the latent variable and a normal distribution around zero. With this the VAE-Encoder infers the posterior distribution:
\begin{align}
\begin{split}
q_\phi(z|s_t, a_{t:t+k-1}) = \mathcal{N}(&\mu_\phi(s_t, a_{t:t+k-1}), \\
&\sigma_\phi^2(s_t, a_{t:t+k-1}))
\end{split}
\end{align}

This latent variable $z$ is then used as input for the VAE-Decoder. At inference time, the latent variable $z$ is set to zero, representing the average style across all demonstrations and producing deterministic predictions.

The KL divergence term $D_{KL}[q_\phi(z|s_t, a_{t:t+k-1}) \| \mathcal{N}(0, I)]$ regularizes the learned posterior distribution toward a standard normal prior. This regularization serves two important functions in ACT. First, it prevents the encoder from learning arbitrarily encodings that overfit to training demonstrations by constraining the geometry of the latent space. Without this term, the encoder could generate highly specific latent encodings for each demonstration that do not generalize to new observations at test time. Second, it ensures that setting $z=0$ at inference time represents a meaningful central tendency of the latent distribution rather than an arbitrary point.

\subsubsection{Transformer Architecture}

The third part of ACT is the integration of a transformer-based model with an encoder-decoder in the VAE-Decoder. This is done to process multimodal inputs, which usually are the joint positions, visual inputs from various cameras, and the latent style variable. These three distinct types of inputs (for the encoder) must be combined into a sequence representation. The latent variable $z \in \mathbb{R}^L$ sampled from the VAE-Encoder provides a compressed representation of the trajectory style, while current proprioceptive information, including joint positions and environmental state variables, is concatenated into a state vector $s_t \in \mathbb{R}^{d_s}$. Visual information from multiple camera viewpoints is processed through pre-trained ImageNet encoders (usually frozen ResNet (\cite{he2016deep}) or ViT (\cite{dosovitskiy2021an}) backbones) to extract spatial feature maps. Each camera image $I_i \in \mathbb{R}^{H \times W \times 3}$ is passed through its backbone network to produce feature maps $F_i \in \mathbb{R}^{H' \times W' \times d}$. These spatial features embeddings are then tokenized by flattening the spatial dimensions (which come from the image), producing a sequence of (visual) tokens $\{f_{i,1}, f_{i,2}, \ldots, f_{i,H'W'}\}$ where each token $f_{i,j} \in \mathbb{R}^d$ represents a spatial patch.

To create a unified input sequence, the encoder concatenates tokens from all modalities. The latent variable $z$ and proprioceptive state $s_t$ are first projected to the model dimension $d$ through learned linear layers, creating tokens $z' \in \mathbb{R}^d$ and $s_t' \in \mathbb{R}^d$. The complete input sequence becomes:
\begin{align}
X = [z'; s_t'; f_{i,1}; f_{i,2}; \ldots, f_{i,H'W'}] \in \mathbb{R}^{S \times d}
\end{align}
where $S = 2 + \sum_i H'W'$ represents the total sequence length across all modalities.

Because the encoder inputs come from heterogeneous modalities, ACT uses distinct positional embeddings for each modality. Non-visual tokens (latent and state) receive learned 1D positional embeddings, while visual tokens from each camera receive 2D sinusoidal positional embeddings that maintain spatial relationships within each feature map.

The transformer encoder processes this multimodal sequence through $N$ identical layers, each containing a multi-head self-attention mechanism followed by a position-wise feedforward network. For input sequence $X^{(l-1)}$ at layer $l-1$, the self-attention computes:
\begin{align}
Q^{(l)} &= X^{(l-1)}W_Q^{(l)} \\
K^{(l)} &= X^{(l-1)}W_K^{(l)} \\
V^{(l)} &= X^{(l-1)}W_V^{(l)} \\
\text{Attention}^{(l)} &= \text{softmax}\left(\frac{Q^{(l)}K^{(l)T}}{\sqrt{d_k}}\right)V^{(l)}
\end{align}

This self-attention mechanism enables the model to integrate information across different modalities, and therefore the resulting encoder output $X^{(N)}$ provides a rich embedding necessary for action prediction, which is done in the transformer decoder.

The transformer decoder generates action sequences through cross-attention to this encoder output. The decoder operates on learned positional embeddings for $k$ action timesteps, initialized as $D_{in} \in \mathbb{R}^{k \times d}$. Each decoder layer performs masked self-attention over the action sequence positions to maintain causal dependencies, followed by cross-attention to the encoder output:

\begin{align}
\text{\small{Self-Attn}} &= \text{\small{Masked-Attn}}(D^{(l-1)}, D^{(l-1)}, D^{(l-1)}) \\
\text{\small{Cross-Attn}} &= \text{\small{Attention}}(\text{\small{Self-Attn}}, X^{(N)}, X^{(N)})
\end{align}

The cross-attention mechanism allows each action prediction position to attend to relevant information across all input modalities processed by the encoder, enabling the decoder to ground action predictions in the appropriate visual and proprioceptive context.

Action sequences are generated by projecting the final decoder output $D^{(L)} \in \mathbb{R}^{k \times d}$ through a learned linear layer:
\begin{equation}
A_{t:t+k-1} = D^{(L)}W_{action} \in \mathbb{R}^{k \times d_a}
\end{equation}
where $d_a$ represents the action dimensionality. The model is trained to minimize L1 loss between predicted and demonstrated action chunks.

During inference, ACT employs temporal ensembling to improve action smoothness. Rather than querying the policy every $k$ timesteps, the policy is queried at every timestep, generating overlapping action chunks. For timestep $t$, multiple predictions may exist from previous policy queries, which are combined through exponential weighting:
\begin{equation}
a_t = \frac{\sum_{i=0}^{k-1} w_i \cdot \hat{a}_{t-i,t}}{\sum_{i=0}^{k-1} w_i}
\end{equation}
where $w_i = \exp(-\beta \cdot i)$ provides exponentially decaying weights favoring more recent predictions, and $\hat{a}_{t-i,t}$ represents the action predicted for timestep $t$ by the policy queried at timestep $t-i$. This temporal ensembling reduces action jitter and provides smoother policy execution while maintaining the benefits of action chunking for training stability.

\subsubsection{Complete Training}

The complete training objective combines reconstruction accuracy of the predicted actions with the KL-Loss of the latent variable:
\begin{align}
\mathcal{L}_{ACT} &= \mathbb{E}_{q_\phi}[\ell(a_{t:t+k-1}, p_\theta(a_{t:t+k-1}|s_t, z))] \nonumber\\
&\quad + \beta \cdot D_{KL}[q_\phi(z|s_t, a_{t:t+k-1}) \| \mathcal{N}(0, I)]
\end{align}

where $\ell(\cdot, \cdot)$ is typically the L1 loss for continuous actions.

The hyperparameter $\beta$ controls the trade-off between reconstruction accuracy and latent regularization. Higher $\beta$ values enforce stronger adherence to the prior distribution, potentially at the cost of reconstruction quality, while lower values allow more flexible latent representations that may capture demonstration-specific variations more precisely. \citet{zhao2023act} set $\beta = 10$, based on empirical validation across multiple manipulation tasks, prioritizing stable latent representations over exact trajectory matching.

\subsubsection{LAV-ACT: Language-Augmented Visual Action Chunking}

\citet{tripathi2025lav} extend ACT by incorporating textual task descriptions alongside visual and proprioceptive inputs, similar to Visual Language Action Models (VLAs) \citep{kim2025openvla}. LAV-ACT modifies the standard ACT backbone through a dual-stream architecture that processes visual and linguistic information separately before fusion.

The architecture replaces the standard ResNet \citep{he2016deep} visual encoder with a multimodal processing pipeline. Visual features are extracted using ResNet, while language descriptions are encoded through a Voltron encoder \citep{karamcheti2023voltron}. These separate embeddings are then fused using Feature-wise Linear Modulation (FiLM) layers \citep{perez2018film}, which allow language features to modulate visual representations through learned affine transformations. The resulting multimodal embeddings serve as inputs to the standard ACT VAE-Decoder, enabling the policy to condition action predictions on both visual observations and linguistic task specifications.

This language conditioning enables task generalization across different verbal instructions while maintaining the temporal consistency benefits of action chunking. However, LAV-ACT still operates solely on low-level sensorimotor data augmented with natural language, without incorporating the structured semantic concepts that ConceptACT leverages for improved sample efficiency.

%TODO Short section about explainable AI and concept methods in general
\subsection{Concept-Based Methods in Explainable AI}

While deep learning models can learn powerful classifiers from low-level features, a major downside is that their predictions are usually not easily human-interpretable, leading to the well-known \textit{"black box"} problem. The field of \textit{Explainable AI} (XAI) developed multiple different methods to make deep networks more understandable, one of which is the so-called \textit{concept-based} methods, which  ground model decisions in high-level concepts.  In this context, a \textit{concept} refers to a human-understandable attribute or property that can be used to describe or categorize inputs. For example, in image classification, concepts might include object properties like \textit{"red color,"} \textit{"round shape,"} or \textit{"metallic texture,"} while in robotics, concepts could represent spatial relationships like \textit{"above,"} \textit{"inside,"} or object states like \textit{"open"} or \textit{"deformable"}. 

The field of concept-based XAI methods can be categorized into two major paradigms:  post-hoc explanation methods and inherently interpretable models. Post-hoc methods, such as \textit{TCAV} (Testing with Concept Activation Vectors) \citep{kim2018interpretability}, analyze trained black-box models to understand how user-defined concepts influence predictions. These analyses are performed at inference time (i.e., post-hoc), without altering the trained model.  

Inherently interpretable models, by contrast, are designed from the ground up to incorporate human-understandable concepts into the training process and thereby make predictions more understandable. One of the most prominent examples in this area is \textit{Concept Bottleneck Models} \citep{koh2020concept}, where the model first predicts a set of concept attributes and then uses these concept predictions to make final task predictions. This concept prediction is performed at a bottleneck, which can later be used to explain the final prediction in terms of the concepts used. Similar approaches are \textit{ProtoPNet} \citep{chen2019looks}, \textit{Self-Explaining Neural Networks (SENN)} \citep{alvarez2018towards}, \textit{Concept Whitening} \citep{chen2020concept}, and \textit{Concept Transformer} \citep{rigotti2022concept}, which is used in this work.

\subsection{Concept Transformers}

\textit{Concept Transformer} by \citeauthor{rigotti2022concept} is also a method for including concepts in the training process, through changes in the transformer architecture. At a high level, the core idea is to modify an attention layer in a defined way so that attention is not on tokens (or patches) but on concepts. This approach addresses a limitation of standard attention-based explanations: while attention weights over input tokens provide some insight into model focus, they operate at a low level that may not correspond to meaningful human concepts.

This change in attention is done through a restructuring of the (multi-head) attention mechanism. As a reminder, in standard attention \citep{vaswani2017attention}, the embeddings for queries (Q), keys (K), and values (V) are all derived from the input sequence through learned linear transformations:

\begin{align}
Q &= XW_Q \\
K &= XW_K \\  
V &= XW_V
\end{align}

where $X \in \mathbb{R}^{P \times d}$ represents the input features (with $P$ patches or tokens and $d$ embedding dimension).

The core idea of Concept Transformers is now to break this uniform calculation of the embeddings by replacing the input (i.e., $X$) in the calculation of $K$ and $V$ with learnable concept embeddings, while the query embeddings ($Q$) are unchanged. Specifically, given a set of $C$ concept embeddings $\mathbf{c}_1, \ldots, \mathbf{c}_C \in \mathbb{R}^d$ organized as $C = [\mathbf{c}_1; \ldots; \mathbf{c}_C] \in \mathbb{R}^{C \times d}$, the attention computation becomes:

\begin{align}
Q &= XW_Q \\
K &= CW_K \\
V &= CW_V
\end{align}

This changes also slightly how the attention weights are computed:
\begin{equation}
\alpha_{sc} = \text{softmax}\left(\frac{QK^T}{\sqrt{d}}\right)_{sc}
\end{equation}

where $s$ indexes input positions ($s = 1, \ldots, S$) and $c$ indexes concepts ($c = 1, \ldots, C$). This produces an attention matrix $A = [\alpha_{sc}] \in \mathbb{R}^{S \times C}$ where each entry represents how much input position $s$ attends to concept $c$.

\subsubsection{Concept Supervision and Training}

The key change in Concept Transformers is that the above-defined cross-attention weights ($\alpha_{pc}$) can be directly used with a supervised loss (alongside human-defined labels) to compute an additional loss term for the model. During training, human experts provide ground-truth concept annotations $H \in \mathbb{R}^{P \times C}$, where $H_{pc} = 1$ indicates that concept $c$ is present and relevant at input position $p$. These labels can then be used to calculate the difference between the predicted concepts ($\alpha_{pc}$) and the given concepts (which usually come from a human expert) with the squared Frobenius norm: 
\begin{equation}
\mathcal{L}_{concept} = \|A - H\|_F^2 = \sum_{p=1}^P \sum_{c=1}^C (\alpha_{pc} - H_{pc})^2
\end{equation}

This supervision ensures that the attention mechanism learns to focus on semantically meaningful concepts rather than arbitrary input patterns.

\subsubsection{Output Computation and Interpretability}

Another key change in the Concept Transformer formulations by \citet{rigotti2022concept} is how the final output of the transformer is computed. Instead of a final MLP layer (which would make the predictions of the model not fully understandable), the final output is a linear combination of the attention weights:

\begin{equation}
\text{output}_p = \sum_{c=1}^C \alpha_{pc} \cdot V_c
\end{equation}

where $V_c$ represents the value vector for concept $c$ and $p$ is each position in the sequence. The advantage of this formulation is that it ensures that the predictions of the model are explicitly constructed from weighted combinations of concept representations, allowing for a direct interpretation of how each concept contributes to the final decision. Therefore, all explanations of the behavior of the Concept Transformer are inherently faithful to the model's decision process.

For the training of the complete model, the standard loss on the task predictions (depending on the current settings) and the concept alignment loss are combined: 
\begin{equation}
\mathcal{L}_{total} = \mathcal{L}_{task} + \lambda \mathcal{L}_{concept}
\end{equation}

where $\lambda$ controls the strength of concept alignment to balance between correct concept predictions and correct task predictions.

\section{Implementation}
\label{sec:implementation}
We extend the standard imitation learning setting to incorporate high-level semantic information that humans naturally use when teaching tasks. The high-level goal is to allow humans to formulate knowledge about the current episode and include this information in the training process.

Crucially, episode-level concept annotations are required only during the demonstration phase. Human demonstrators naturally observe object properties and task constraints while performing demonstrations, making episode-level annotation straightforward and non-intrusive. Once training is complete, the resulting policy operates on normal inputs without requiring any concepts at deployment time.

\subsection{Problem Formulation with Concepts}

In addition to the demonstration dataset $\mathcal{D} = \{(s_i, a_i)\}_{i=1}^N$ of state-action pairs, we assume access to concept annotations produced by humans during the demonstration phase. Rather than creating a single flat concept vector, we maintain separate concept annotations for each concept class, better reflecting the semantic structure of the problem.

We define a set of concept classes $\mathcal{T} = \{T_1, T_2, \ldots, T_K\}$ which represent higher-level concept groups, such as object colors, shapes, or spatial relationships. Each concept class $T_j$ contains a finite set of mutually exclusive concept values representing the concrete instantiation of that concept in the current episode. For example, a color concept class contains $T_{\text{color}} = \{\text{red}, \text{green}, \text{blue}, \text{yellow}\}$ as potential concept values.

For each episode $e$ and concept class $T_j$, we define:
\begin{equation}
    c^{(e), T_j} \in \{0,1\}^{|T_j|}
\end{equation}

where $c^{(e), T_j}$ is a one-hot encoding indicating which specific concept value within concept class $T_j$ is present in episode $e$. The complete concept annotation for episode $e$ is thus represented as the set:
\begin{equation}
    \mathcal{C}^{(e)} = \{c^{(e), T_1}, c^{(e), T_2}, \ldots, c^{(e), T_K}\}
\end{equation}

\textbf{Example:} 
A episode for a manipulation tasks contains a blue objects, with a rough surface (important for gripping) and it should be dropped in "zone B", with the following concept classes:  $T_{\text{color}} = \{\text{red}, \text{green}, \text{blue}, \text{yellow}\}$, $T_{\text{shape}} = \{\text{cube}, \text{cylinder}, \text{sphere}\}$, $T_{\text{surface}} = \{\text{smooth}, \text{rough}\}$, and $T_{\text{zone}} = \{\text{A}, \text{B}, \text{C}\}$, the annotations would be:
\begin{align*}
    c^{(e), T_{\text{color}}} &= [0, 0, 1, 0] \\
    c^{(e), T_{\text{shape}}} &= [1, 0, 0] \\
    c^{(e), T_{\text{surface}}} &= [0, 1] \\
    c^{(e), T_{\text{zone}}} &= [0, 1, 0]
\end{align*}

\subsubsection{Episode-to-Step Concept Mapping}

Since policy training operates at the step level while human demonstrators provide concept annotations at the episode level, we require a mapping function that projects episode concepts onto individual timesteps. For each concept class $T_j$, we define:
\begin{equation}
    \phi_j: \mathcal{C}^{(e)} \times \mathcal{E} \rightarrow \{0,1\}^{|T_j|}
\end{equation}
where $\mathcal{E}$ represents the set of all episodes, such that $\phi_j(\mathcal{C}^{(e)}, e) = c^{(e), T_j}$ for all steps within episode $e$.

This constant mapping assumes that relevant semantic concepts remain consistent throughout an episode. Since humans assign concepts only at the episode level, this assumption holds by design. For each training step $t$ belonging to episode $e(t)$, the concept vector for class $T_j$ becomes:
\begin{equation}
    c_t^{T_j} = \phi_j(\mathcal{C}^{(e(t))}, e(t)) = c^{(e(t)), T_j}
\end{equation}

This class-wise projection of episode-level concepts to timestep-level enables clean integration with both our prediction head and concept transformer approaches, eliminating the need for concatenation and subsequent splitting of concept vectors.

\subsection{ConceptACT Architecture}

To incorporate concept learning into ACT training, we modify the architecture by replacing the final layer of the transformer encoder within the VAE-Decoder component. All other components of the ACT architecture, including the VAE-Encoder, transformer decoder, and action prediction head, remain unchanged, ensuring that our modifications are minimal and focused.

Our decision to modify the transformer encoder (within the VAE decoder) is motivated by several architectural considerations. The VAE-Encoder processes only proprioceptive joint information and action sequences to maintain computational efficiency, excluding visual inputs that are essential for learning image-based concepts, such as object colors and shapes. The transformer decoder, while powerful for sequential action prediction, typically consists of only a single layer and is specifically designed for the generative task of producing action sequences, making it less suitable for the discriminative concept classification task.

In contrast, the transformer encoder within the VAE-Decoder receives the complete multimodal input, including visual features from multiple camera viewpoints, proprioceptive state information, and latent variables, providing the rich representational foundation necessary for concept learning. By positioning concept learning at the final encoder layer, we ensure that concept information is learned from the highest-level multimodal representations while still being able to influence the features passed to the action decoder.

A core feature (or constraint) of our contributions is that our concept annotations only change the learning phase of the ACT framework. During inference (i.e., deployment), no knowledge of concepts is needed to use the policy to predict actions. This constraint (or strength) limits the type of possible concept integrations, resulting in an auxiliary loss, which is a straightforward solution to this constraint.

\subsubsection{Class-Aware Concept Transformer Integration}

We replace the final layer of the transformer encoder with a class-aware concept transformer layer. This modification differs from standard concept transformer architectures by recognizing that different concept types possess distinct semantic structures and cardinalities that should be processed through dedicated attention mechanisms.

Rather than treating all concepts uniformly as in standard concept transformer approaches, our class-aware architecture maintains separate learnable concept embeddings for each concept class $T_j$. For a concept class $T_j$ with cardinality $|T_j|$ (representing the number of possible values within that class), we define concept embeddings:

\begin{equation}
E_j \in \mathbb{R}^{|T_j| \times d}
\end{equation}

where $d$ is the embedding dimension and each row represents a learnable embedding for one concept value within class $T_j$.

Given encoder input $X \in \mathbb{R}^{S \times d}$ where $S$ represents the sequence length (encompassing tokens from images, proprioceptive state, and other modalities), each concept class $T_j$ computes its own attention mechanism through class-specific projection matrices:

\begin{align}
Q_j &= XW_Q^{(j)} \\
K_j &= E_j W_K^{(j)} \\
V_j &= E_j W_V^{(j)}
\end{align}

where $W_Q^{(j)}, W_K^{(j)}, W_V^{(j)} \in \mathbb{R}^{d \times d}$ are learned projection matrices specific to concept class $T_j$, and $E_j \in \mathbb{R}^{|T_j| \times d}$ are the learnable concept embeddings for class $T_j$.

The attention weights for concept class $T_j$ are computed as:

\begin{equation}
\alpha_{sc}^{(j)} = \text{softmax}\left(\frac{Q_j K_j^T}{\sqrt{d}}\right)_{sc}
\end{equation}

where the subscript $s$ indexes sequence positions ($s = 1, \ldots, S$) and $c$ indexes the concepts within class $T_j$ ($c = 1, \ldots, |T_j|$). Due to the specific dimensionality of concept embeddings $E_j$, the attention weight matrix $\alpha^{(j)}$ has shape $S \times |T_j|$ rather than the typical sequence length by embedding dimension. This structure reflects our class-aware design where each sequence position attends to exactly $|T_j|$ concept options within the specific class.

\subsubsection{Concept Transformer Output Computation}

For each sequence position $s$ and concept class $T_j$, the concept transformer produces an output by computing the attention-weighted combination of concept value vectors:

\begin{equation}
\text{o}_s^{(j)} = \sum_{c=1}^{|T_j|} \alpha_{sc}^{(j)} V_{jc}
\end{equation}

where $V_{jc} \in \mathbb{R}^d$ represents the $c$-th value vector (corresponding to the $c$-th concept within class $T_j$). This operation yields $\text{o}_s^{(j)} \in \mathbb{R}^d$ for each sequence position $s$.

To integrate concept information with the original sequence representation, we concatenate the concept outputs from all classes with the original input for each sequence position:

\begin{equation}
X_s^{\text{enhanced}} = [X_s; \text{o}_s^{(1)}; \text{o}_s^{(2)}; \ldots; \text{o}_s^{(K)}]
\end{equation}
%TODO note that this is different from the original concept transformer. Where the ouput is purely a linear weighingt of concept
where $X_s^{\text{enhanced}} \in \mathbb{R}^{d \cdot (K+1)}$ represents the enhanced representation for sequence position $s$ that combines original features with concept-aware representations from all $K$ concept classes. This is a slight modification from the original concept transformer, where the is purely computed from a weighted sum of concepts. Here, we introduce a residual connection $X_s$, which allows the decoder upstream to theoretically ignore the output of the concept transformer. 

A projection layer then maps this concatenated representation back to the original embedding dimension:

\begin{equation}
Y_s = W_{\text{proj}} X_s^{\text{enhanced}} + b_{\text{proj}}
\end{equation}

where $W_{\text{proj}} \in \mathbb{R}^{d \times d(K+1)}$ and $b_{\text{proj}} \in \mathbb{R}^d$ are learned parameters. The resulting sequence $Y = [Y_1; Y_2; \ldots; Y_S] \in \mathbb{R}^{S \times d}$ maintains the original dimensionality and serves as input to the ACT decoder. 

\subsubsection{Concept Prediction \& Loss formulation}
While the sequence $Y$ serves as the major output of the concept transformer, the more crucial change of our architecture is the usage of the auxiliary concept attention $\alpha_{sc}^{(j)}$, which is used further to align the concept transformer with the human concepts. 
The encoder input sequence $X$ contains tokens representing various modalities that collectively describe a single timestep in the demonstration trajectory. Although we process this information as a sequence of tokens for computational efficiency, it conceptually represents a unified observation at one demonstration step. Therefore, we must aggregate across the sequence dimension to produce a single concept prediction per class for each demonstration step. \footnote{Please note that sequence here means in terms of patches/token extracted from the input states \& images. Not a sequence in the demonstration dataset.}

We obtain concept predictions for class $T_j$ through mean pooling across sequence positions:

\begin{equation}
\hat{c}^{T_j} = \frac{1}{S}\sum_{s=1}^S \alpha_{sc}^{(j)}
\end{equation}

This mean pooling operation transforms the sequence-level concept representations into a single prediction vector for each concept class, ensuring that our final concept predictions correspond to the episode-level annotations provided by human demonstrators. 

Our concept loss combines cross-entropy losses across all concept classes, incorporating label smoothing as a regularization technique. For each concept class $T_j$, we compute the cross-entropy loss between the predicted logits and the ground-truth one-hot concept vector $c^{T_j}$.

For a concept class $T_j$ with ground-truth concept vector $c^{T_j} \in \{0,1\}^{|T_j|}$ and label smoothing parameter $\epsilon$, the smoothed target distribution becomes:
\begin{equation}
q_k^{T_j} = \begin{cases}
1-\epsilon+\frac{\epsilon}{|T_j|} & \text{if } c_k^{T_j} = 1 \\
\frac{\epsilon}{|T_j|} & \text{if } c_k^{T_j} = 0
\end{cases}
\end{equation}

The complete concept loss combines smoothed cross-entropy losses across all concept classes:
\begin{equation}
\mathcal{L}_{\text{concept}}^{\text{CE}} = \sum_{j=1}^K \mathbb{E}\left[-\sum_{k=1}^{|T_j|} q_k^{T_j} \log \text{softmax}(\hat{c}^{T_j})\right]
\end{equation}

where $\hat{c}^{T_j} $ represents the predicted logits for concept class $T_j$, obtained by passing the mean-pooled concept output as described in the previous section.

\subsubsection{Complete Training Objective}

The final ConceptACT training objective integrates the class-aware concept loss with the standard ACT components:

\begin{equation}
\mathcal{L}_{\text{total}} = \mathcal{L}_{ACT} + \lambda_{\text{concept}} \mathcal{L}_{\text{concept}}^{\text{CE}}
\end{equation}

where $\mathcal{L}_{ACT}$ encompasses both action reconstruction loss and KL divergence regularization as defined in the standard ACT formulation.

%\subsubsection{Architectural Properties}

This class-aware concept transformer layer replaces the final encoder layer within the VAE-Decoder component, positioning concept learning at the highest representational level before action prediction. The architecture addresses three specific limitations of standard concept transformers in our setting. First, the separate attention mechanisms per concept class eliminate competition between semantically unrelated concepts (e.g., colors competing with shapes for attention), allowing each class to develop specialized representations. Second, the variable-dimension concept embeddings $E_j \in \mathbb{R}^{|T_j| \times d}$ accommodate concept classes with different cardinalities without requiring padding or masking operations that could interfere with learning. Third, the cross-entropy loss per class enforces the constraint that exactly one concept per class should be active, matching the one-hot structure of our human annotations and preventing the model from learning spurious multi-label predictions within a single concept class.

After training is complete, the concept prediction components can be discarded during inference, allowing the policy to be used normally for action prediction. This ensures that concept learning aids training without affecting deployment efficiency.

\section{Evaluation}\label{sec4}

While ConceptACT is generally applicable to any imitation learning domain, we evaluate our approach in a robotic manipulation setting where concept-based guidance provides the most significant benefit. Robotic tasks naturally exhibit complex state-action relationships that benefit from semantic reasoning, while their visual and proprioceptive observations enable intuitive concept definitions around object properties and spatial relationships.

We structure our evaluation around two complementary research questions that address the core bottlenecks in practical imitation learning: (1) \textit{Sample efficiency}: Can concept supervision reduce the number of human demonstrations required to achieve target performance? (2) \textit{Learning efficiency}: Can concept supervision accelerate convergence with fixed demonstration data? These questions correspond to the two primary cost factors in deploying imitation learning systems: human time for data collection and computational resources for training.

To address these questions systematically, we design two manipulation tasks that require different types of concept reasoning while maintaining consistent annotation methodology. Our experimental approach isolates the effect of concept integration by comparing ConceptACT variants against standard ACT across multiple data regimes and training horizons.

\subsection{Hardware Configuration}

Our experimental setup utilizes a bilateral robotic system comprising two identical SO-100 robotic arms \citep{SO-100}. One arm serves as the leader for human demonstration and teleoperation, while the other functions as the follower for policy execution and evaluation. This leader-follower configuration enables intuitive data collection: human demonstrators directly manipulate the leader's arm through kinesthetic teaching, with joint positions and trajectories recorded in real-time. 

The system is equipped with two cameras, providing complementary viewpoints: a gripper camera attached to the end-effector of the follower arm for detailed manipulation views, and a scene camera positioned to capture the entire workspace from a fixed overhead perspective. This dual-camera setup ensures comprehensive visual coverage of both fine-grained manipulation details and global scene context.

\begin{figure}[h]
    \centering
    \begin{minipage}[t]{0.5\linewidth}
        \centering
        \begin{subfigure}[t]{\textwidth}
            \centering
            \includegraphics[width=\textwidth,keepaspectratio]{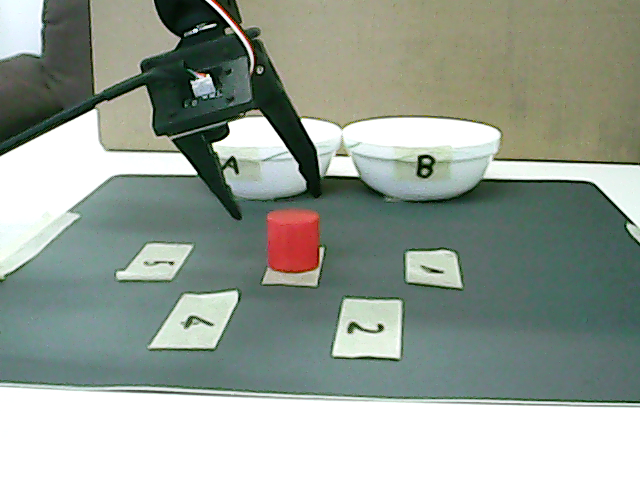}
            \caption{Scene Camera}
            \label{fig:scene}
        \end{subfigure}
        \\[0.1cm]
        \begin{subfigure}[t]{\textwidth}
            \centering
            \includegraphics[width=\textwidth,keepaspectratio]{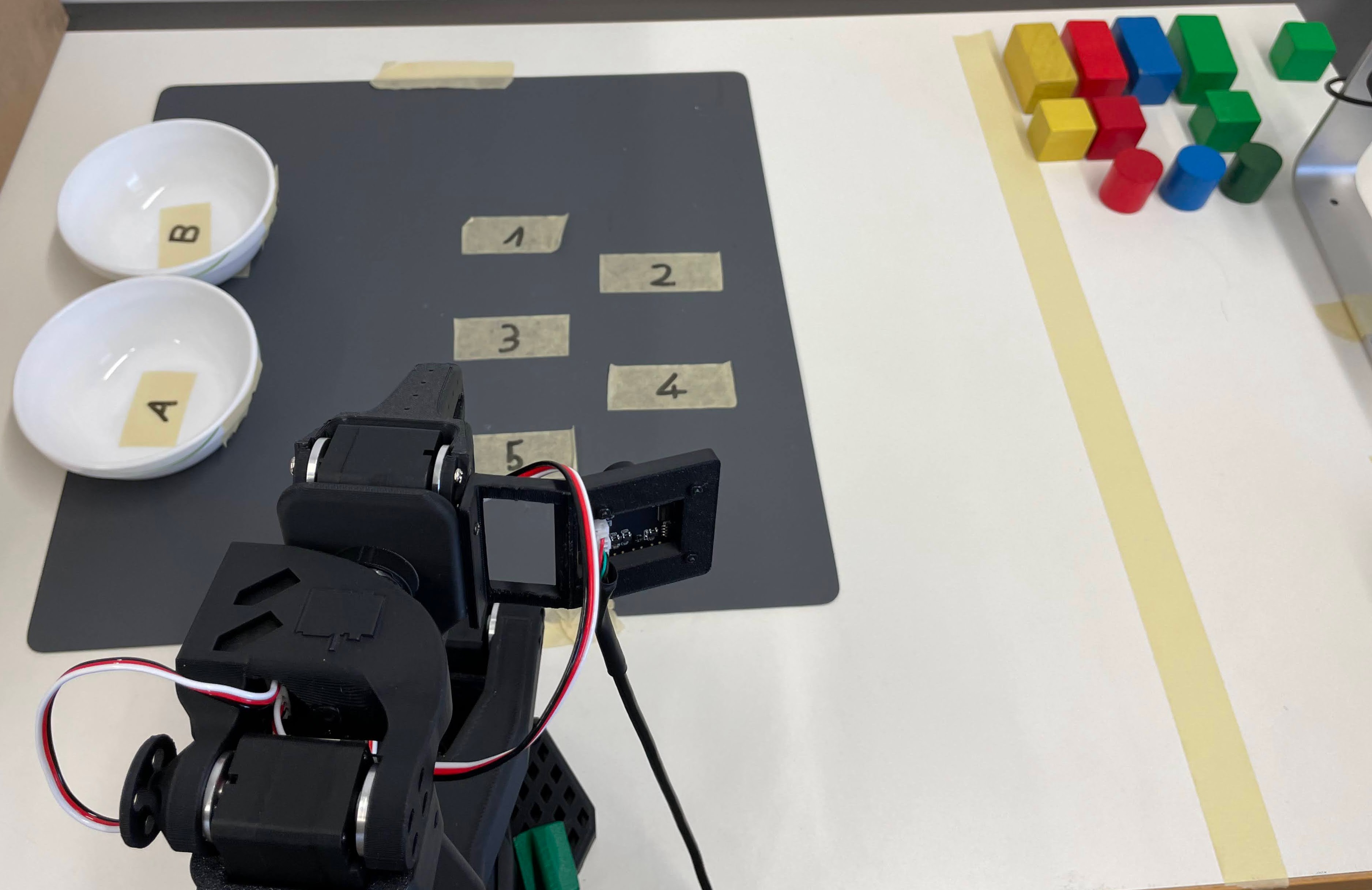}
            \caption{Experimental setup}
            \label{fig:setup}
        \end{subfigure}
    \end{minipage}
    \hfill
    \raisebox{-0.5cm}{%
        \begin{subfigure}[c]{0.45\linewidth}
            \centering
            \includegraphics[width=\textwidth,keepaspectratio]{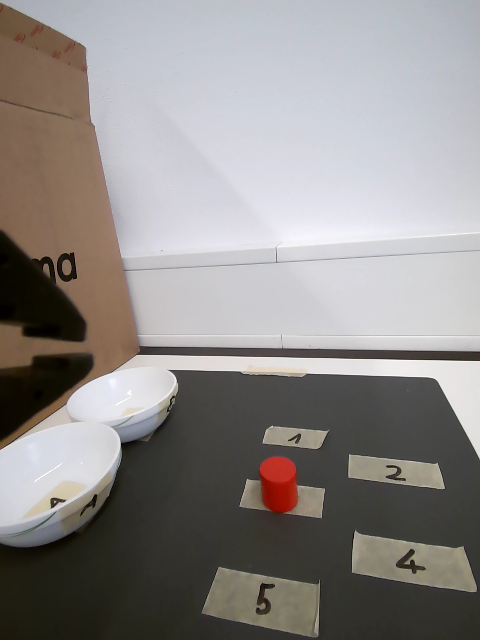}
            \caption{Wrist Camera}
            \label{fig:wrist}
        \end{subfigure}
    }
    \caption{Experimental hardware configuration showing different camera viewpoints and the bilateral robotic system}
    \label{fig:experimental_setup}
\end{figure}
\subsection{Task Description and Concept Annotations}

We evaluate ConceptACT on two robotic manipulation tasks that require logical reasoning beyond simple pick-and-place motions. Both tasks share similar motion primitives but differ in their underlying constraint structures, validating the generality of our approach across manipulation scenarios.

\textbf{Task 1 (Sorting):} The robot must grasp objects of varying shapes (cube, cylinder, rectangular prism) and colors (red, green, blue, yellow) from randomized positions and sort them into two collection areas according to conditional rules combining both attributes. The sorting rule creates a logical disjunction where objects are directed to Area A if they satisfy specific shape-color combinations, with all other objects sorted to Area B (see Appendix~\ref{appendix:task1} for complete task specification and sorting rules).

\textbf{Task 2 (Ordering):} Sequential placement of three objects into a single collection area according to multi-constraint optimization rules. The robot must satisfy simultaneous constraints on object cardinality, positional placement, color hierarchy, and adjacency relationships (see Appendix~\ref{appendix:task2} for complete constraint system and longer task description).

For Task 1, we employ a hierarchical scoring system (0-3) capturing pick success, placement success, and sorting correctness. Task 2 uses a similar metric (0-6) scoring correct sequential placement. We evaluate on held-out test episodes across 10 different trained models (different by random seeds).

Human demonstrators collected episodes at 50Hz, recording joint positions, camera feeds, and timestamped actions. Critically, each episode was annotated with concept vectors as one-hot encodings: for Task 1, shape ($c^{\text{shape}} \in \{0,1\}^3$), color ($c^{\text{color}} \in \{0,1\}^4$), and pickup location ($c^{\text{location}} \in \{0,1\}^5$); for Task 2, shape and color for each of three objects plus ordering sequence. Demonstrators provided these annotations during data collection by identifying object properties at the initiation of each episode.

For each plot and metric (Probablity of Improvement and Optimality Gap) we calculate the mean and confidence intervals. For the metrics, these are computed via stratified bootstrapping (see \cite{agarwal2021deep}). 

\subsubsection{Baseline Methods and Implementation Details}

We compare five methods to isolate different aspects of concept integration:

\begin{itemize}
\item \textbf{ACT}: Standard baseline without concept information
\item \textbf{ConceptACT-Transformer}: Full concept transformer integration (our primary method)
\item \textbf{ConceptACT-Heads}: Architectural ablation using simple prediction heads (see Appendix~\ref{appendix:prediction_heads} for a detailed description). The goal of this ablation is to show that not only is the information about concepts important, but also our integration through the Concept Transformer is crucial.
\item \textbf{LAV-ACT - Specific}: The LAV-ACT architecture receives task descriptions during both training and inference. In the "Specific" configuration, the model receives detailed task descriptions that contain information about the current scenario's object properties and constraints, which our ConceptACT methods provide only during training.
\item \textbf{LAV-ACT - Generic}: Since detailed task descriptions at inference time represent privileged information unavailable in practical deployments, we test LAV-ACT with generic task descriptions that leave out specific object properties. For example, we replace "Pick up the red cube and place it into zone A" with "Pick up the object and place it in the desired zone." This configuration provides a more realistic comparison, as detailed semantic descriptions are costly to obtain at deployment time compared to the one-time annotation cost during demonstration collection.
% TODO Link and we need that section in the first chapter somewhere that concepts are easier to annotate at training time.
\end{itemize}

A key practical distinction is that ConceptACT methods use concepts only during training, while LAV-ACT requires language descriptions at deployment time. This difference has significant implications for real-world applicability.

All models use identical hyperparameters, which where collected from the reference literature or implementation. For a full list of hyperparameter please see the appendix \ref{appendix:hyperparams}.

\subsection{Sample Efficiency Experiments}

We evaluate whether concept supervision reduces the number of human demonstrations required to achieve target performance. Using both manipulation tasks, we train models on varying dataset sizes (33\%, 50\%, 66\%, 83\%, and 100\% of available demonstrations) and measure real robot performance on 10 held-out test episodes. Each configuration uses 10 random seeds to ensure statistical reliability.

\subsubsection{Task 1: Sorting with Limited Demonstrations}

\begin{figure}
    \centering
    \includegraphics[width=\linewidth]{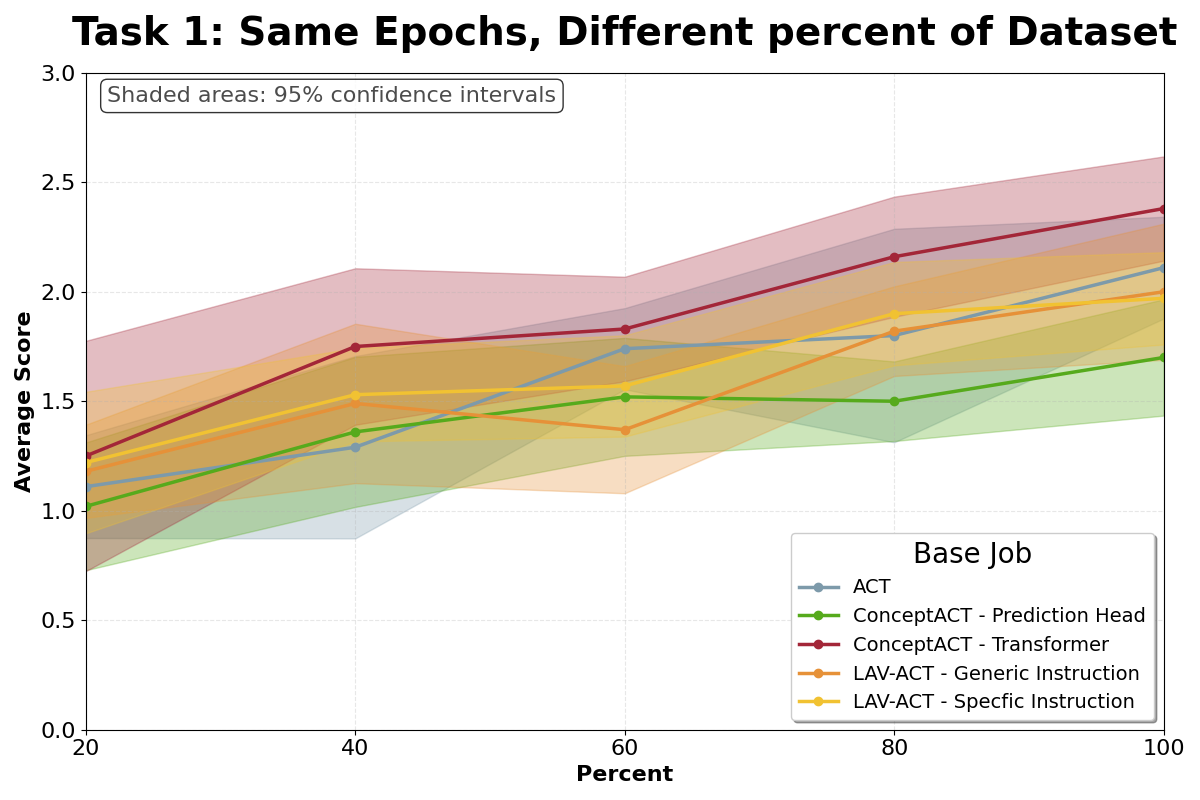}
    \caption{Sample efficiency results on task 1. Real robot evaluation performance is measured on different sizes of the dataset (with respect to the maximum size) on 10 test episodes (higher is better). Each method was trained with 10 random seeds; bold lines represent means, shaded areas show variance.}
    \label{fig:task1_sample_eff}
\end{figure}

Figure~\ref{fig:task1_sample_eff} demonstrates that ConceptACT-Transformer achieves superior performance across all dataset sizes, with particularly pronounced advantages in low-data regimes. The method reaches near-optimal performance with only 50\% of demonstrations, while standard ACT requires the full dataset to achieve comparable results.

% Insert Plot with Learning Progress

\subsubsection{Task 2: Ordering with Limited Demonstrations}

\begin{figure}
    \centering
    \includegraphics[width=\linewidth]{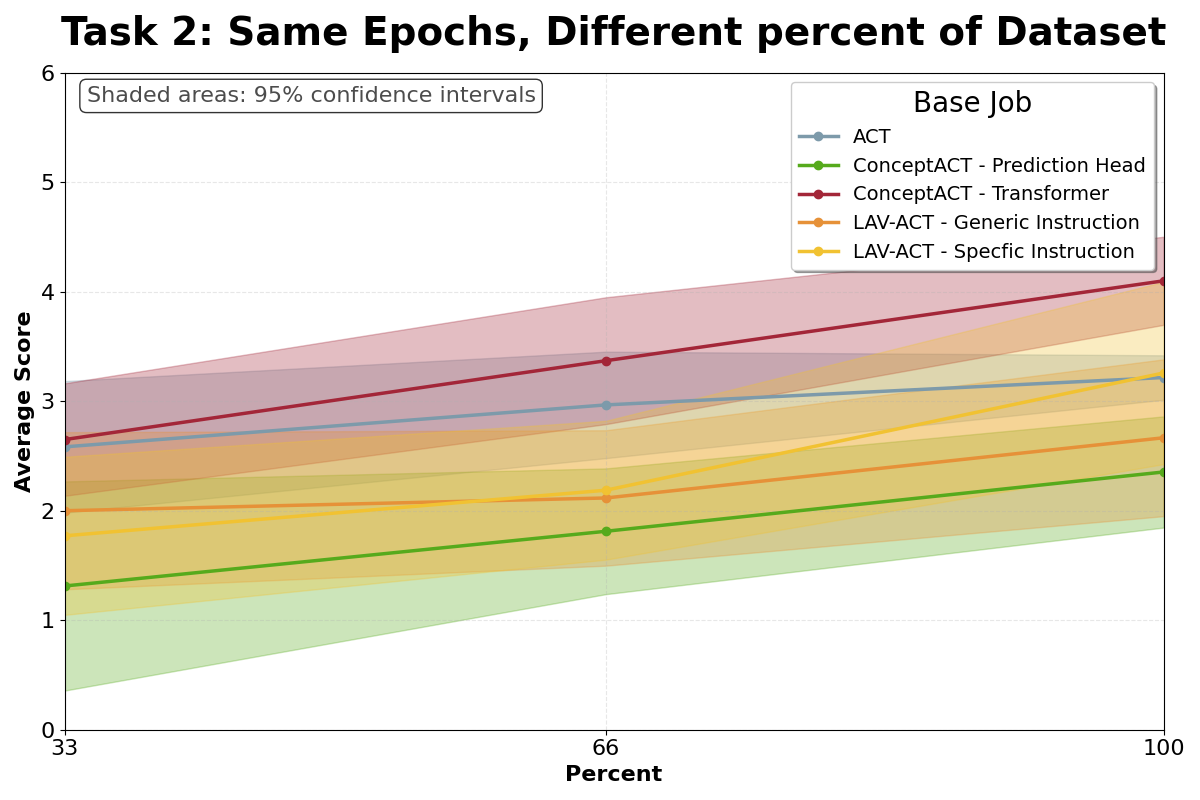}
    \caption{Sample efficiency results on task 1. Real robot evaluation performance is measured on different sizes of the dataset (with respect to the maximum size) on 10 test episodes (higher is better). Each method was trained with 10 random seeds; bold lines represent the means, and shaded areas show the variance.}
    \label{fig:task2_sample_efficiency}
\end{figure}

The ordering task amplifies the benefits of concept guidance due to its increased conceptual complexity, which requires simultaneous reasoning about multiple objects and sequential constraints. As shown in Figure~\ref{fig:task2_sample_efficiency}, ConceptACT-Transformer maintains consistent performance advantages across all data regimes, while the prediction head ablation (ConceptACT-Heads) performs poorly, particularly degrading as concept complexity increases.

%Insert Plot with Learning Progress
% Table with Probability of Improvement (calculated for tasks 1 & 2 and both together. 

The overlapping confidence intervals in both plots necessitate statistical analysis to quantify performance differences. Table~\ref{tab:poi} summarizes results using Probability of Improvement calculated via stratified bootstrapping. ConceptACT-Transformer achieves statistically significant performance gains across both tasks, with robust improvements over the architectural ablation. Notably, our approach outperforms both LAV-ACT variants despite using fewer parameters (see \ref{appendix:parameters}), demonstrating that architectural integration of concepts provides superior sample efficiency compared to language conditioning approaches. Importantly, ConceptACT achieves these performance improvements without requiring any additional information beyond standard sensorimotor inputs at the time of deployment.

The poor performance of ConceptACT-Heads, especially on the concept-rich Task 2, validates our architectural design choice. Simply adding concept prediction losses without modifying attention mechanisms fails to provide meaningful benefits and can even hinder performance when concept complexity increases.

\begin{table*}[h]
    \centering
    \caption{Sample efficiency results: Probability of Improvement aggregated across full training trajectory (higher is better). Confidence intervals computed via bootstrapping. Bold numbers indicate results which can be deemed statistically significant.}\label{tab:poi}%
    \begin{tabular}{l|cl|cl|cl}
    \toprule
    Method    & \multicolumn{5}{|c}{Probability of Improvement} & [95\% CI]\\
    \midrule
    ConceptACT - Transformer    & \multicolumn{2}{c}{Task 1} & \multicolumn{2}{c}{Task 2} &\multicolumn{2}{|c}{Both}\\
    \midrule    
    \multicolumn{1}{r|}{$>$ ACT}    & 0.64 & [0.53, 0.75] & 0.70 & [0.57, 0.83] & 0.66 & [0.58, 0.75]\\
    \multicolumn{1}{r|}{$>$ LAV-ACT - Specifc}     & 0.65 & [0.54, 0.76] & 0.81 & [0.67, 0.91] & 0.71 & [0.63, 0.79\\
    \multicolumn{1}{r|}{$>$ LAV-ACT - Generic}     & 0.71 & [0.60, 0.81] & 0.83 & [0.72, 0.92] & 0.75 & [0.72, 0.83]\\
    \multicolumn{1}{r|}{$>$ ConceptACT - Heads}     & 0.79 & [0.69, 0.87] & 0.93 & [0.83, 1.00] & 0.84 & [0.77, 0.90]\\
    \midrule
    ConceptACT - Heads    & \multicolumn{2}{c}{Task 1} & \multicolumn{2}{c}{Task 2} &\multicolumn{2}{|c}{Both}\\
    \midrule
    \multicolumn{1}{r|}{$>$ ACT}    & 0.31 & [0.21, 0.43] & 0.13 & [0.04, 0.25] & 0.24 & [0.17, 0.32]\\
    \bottomrule
    \end{tabular}
\end{table*}

\subsection{Learning Efficiency Experiment}

To address our second research question, whether concept supervision accelerates convergence with fixed demonstration data, we examine training dynamics across the full 10,000-step training horizon using 100\% of available demonstrations. This analysis focuses on computational efficiency: can concept learning reduce the training time required to achieve target performance levels?

Using the sorting task (Task 1), we track both test loss convergence and robot evaluation performance throughout training, measuring how quickly each method approaches its final performance. This experiment isolates the computational benefits of concept supervision from sample efficiency gains, addressing scenarios where demonstration data is abundant but training time remains a constraint.

\begin{figure*}[h]
    \centering
    \begin{subfigure}[t]{0.48\textwidth}
        \centering
        \includegraphics[width=\textwidth]{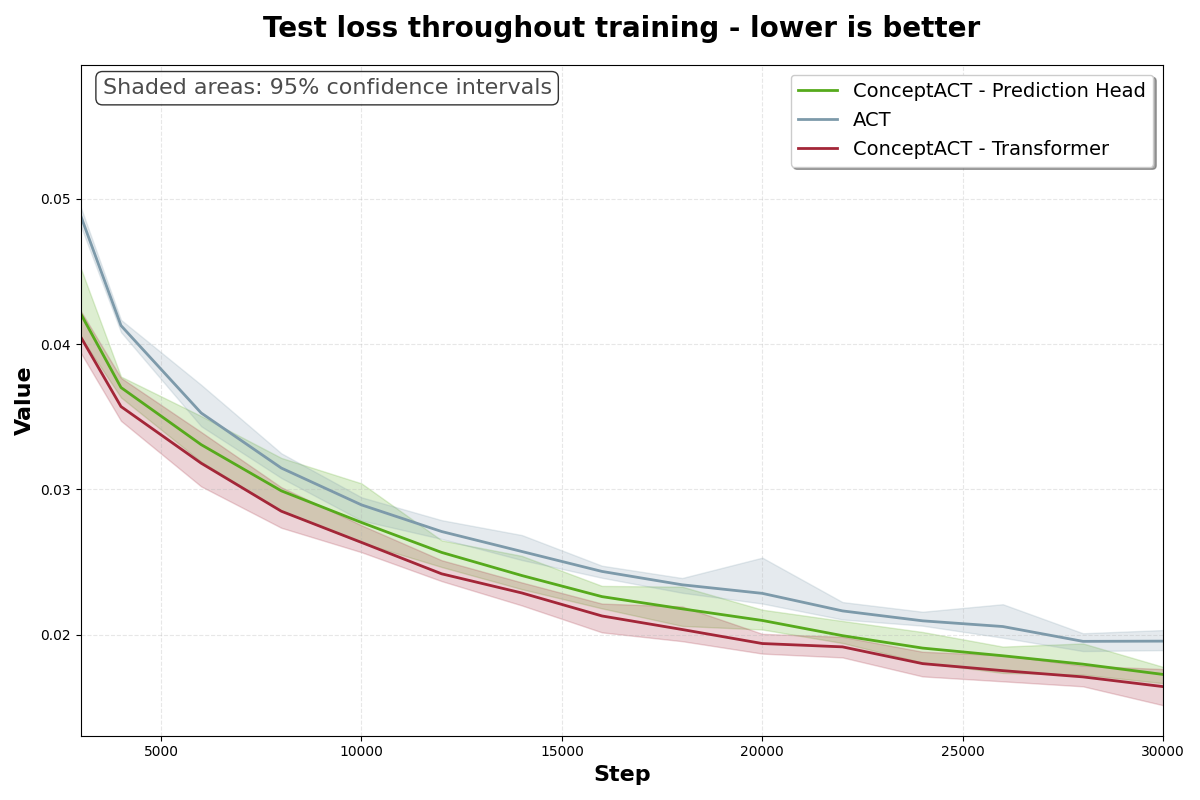}
        \caption{Test loss during training}
        \label{fig:test_loss}
    \end{subfigure}
    \hfill
    \begin{subfigure}[t]{0.48\textwidth}
        \centering
        \includegraphics[width=\textwidth]{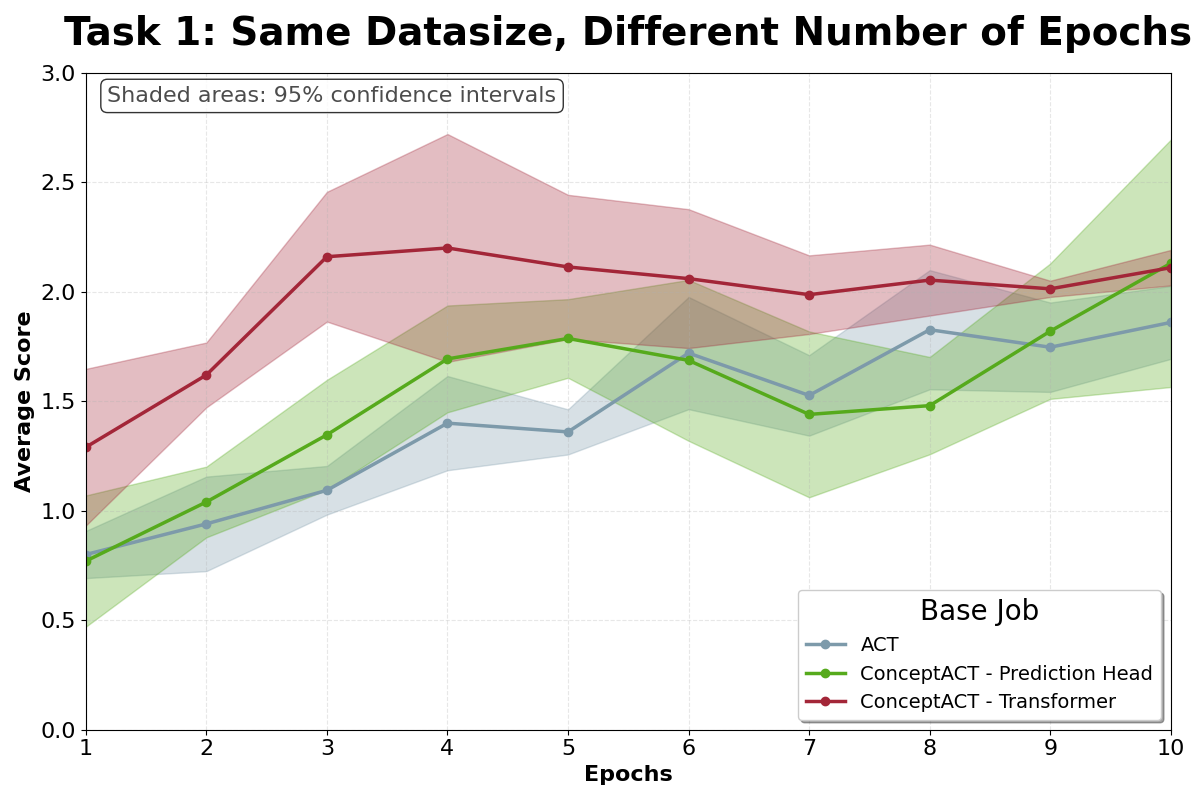}
        \caption{Robot evaluation performance}
        \label{fig:robot_eval}
    \end{subfigure}
    \caption{Learning efficiency results comparing convergence speed across methods. (a) Test loss on 20\% holdout set throughout training (lower is better). (b) Real robot evaluation performance is measured every 3,000 training steps on 10 test episodes (higher is better). Each method was trained with 5 random seeds; bold lines represent means, shaded areas show variance.}\label{fig:results}
\end{figure*}

As shown in Figure~\ref{fig:results}, ConceptACT with Concept Transformer integration demonstrates substantially faster convergence during early training phases. This acceleration is evident in both test loss metrics (Figure~\ref{fig:results}a) and real robot evaluation performance (Figure~\ref{fig:results}b). The Concept Transformer method reaches its final performance approximately 6,000 steps earlier than standard ACT, representing a 40\% reduction in training time to achieve comparable results.

The performance gap between methods becomes less pronounced in later training phases, which we attribute to the relative simplicity of the current task given the available demonstration data. Nevertheless, ConceptACT's faster convergence provides substantial benefits in scenarios where training time constitutes a significant computational cost or when rapid deployment is required.

The prediction head variant again yields minimal improvement over standard ACT, reinforcing our finding that the architectural integration method critically determines the effectiveness of concept supervision. This consistent pattern across both sample efficiency and learning efficiency experiments demonstrates that naive concept inclusion provides negligible benefits regardless of data regime.

Qualitatively, we observed distinct failure modes that reveal the underlying learning differences. Standard ACT policies frequently exhibited boundary confusion, dropping objects precisely at the border between collection areas or balancing them on edges, indicating successful manipulation learning but failure to internalize the sorting logic. ConceptACT policies never displayed this behavior, suggesting more robust acquisition of the underlying task structure encoded in concept annotations.

To quantify these learning efficiency gains, we computed optimality gaps across the full training trajectory and report aggregate performance with 95\% confidence intervals using bootstrapping. Table~\ref{tab:optimality} demonstrates that while both concept-based methods reduce the optimality gap, only the Concept Transformer integration achieves statistically significant improvement over baseline ACT.

\begin{table}[h]
\caption{Learning efficiency results: Optimality gap aggregated across full training trajectory (lower is better). Confidence intervals computed via bootstrapping across 5 random seeds.}\label{tab:optimality}%
\begin{tabular}{@{}lcc@{}}
\toprule
Method & Optimality Gap & 95\% CI \\
\midrule
ACT & 0.549 & [0.48, 0.58]\\
ConceptACT - Heads & 0.515 & [0.45, 0.56]\\
ConceptACT - Transformer & 0.317 & [0.27, 0.40]\\
\bottomrule
\end{tabular}
\end{table}

These results demonstrate that concept supervision provides computational efficiency benefits complementary to sample efficiency gains. The 42\% reduction in optimality gap for ConceptACT-Transformer translates to meaningful training time savings in practical deployment scenarios where computational resources constrain model development cycles.

\section{Related Work}\label{sec5}

Our work integrates high-level semantic concepts into imitation learning to improve sample efficiency and interpretability. Here we review related approaches that incorporate auxiliary information, particularly concepts, into various learning paradigms.

\citet{hristov2021learning} label demonstrated trajectories with high-level spatial concepts ("behind", "on top") and temporal concepts ("quickly"). Their method learns disentangled representations that separate task-relevant features from task-irrelevant variations. While they focus on trajectory-level semantic labeling similar to our episode-level concepts, their approach targets explainability through disentanglement rather than improving sample efficiency through auxiliary supervision.

\citet{cubek2015high} employ conceptual spaces theory for learning from demonstration, using subspace clustering to identify relevant conceptual dimensions from sensory data. Their method automatically discovers conceptual representations that bridge low-level sensory input and high-level task understanding but relies on hard-coded high-level actions and leverages symbolic planning. Unlike their unsupervised concept discovery, ConceptACT uses human-provided concept annotations during training to guide the learning process explicitly.

\citet{stepputtis2019language} combine natural language, vision, and motion for abstract task representation in imitation learning. Their multi-modal approach learns policies conditioned on language instructions, enabling generalization across task variations. While both approaches integrate high-level semantic information, their language conditioning operates at inference time for task specification, whereas ConceptACT uses concept annotations during training to improve learning efficiency.

\citet{ye2025licorice} reduces concept annotation requirements in RL to 500-5000 labels through active learning and concept ensembles. They interleave concept learning with online RL training, where the agent explores and learns simultaneously. In contrast, ConceptACT operates in offline imitation learning with fixed demonstration datasets and episode-level annotations. We integrate concepts directly into the transformer architecture through attention mechanisms, whereas theirs maintains separate concept and policy networks.

\citet{das2023state2explanation} shows that generating concept-based explanations during training improves learning through the Protégé Effect. Their agents learn to produce natural language explanations grounded in predefined concepts as auxiliary outputs. ConceptACT instead uses human-provided concepts as auxiliary inputs, modifying the transformer's internal attention to incorporate semantic information directly into action prediction rather than generating separate explanations.

\cite{haramatientity} decomposes visual observations into entity-specific representations, enabling agents trained with 3 objects to generalize to 10+ objects. They automatically extract object-level structure through architectural inductive biases without human supervision. ConceptACT differs by incorporating human-defined semantic concepts at the episode level, explicitly leveraging domain knowledge about task structure rather than discovering it automatically.

The idea of restricting the network structures to concepts has been quite successfully investigated in supervised learning. Besides Concept Transformer \citep{rigotti2022concept}, other notable approaches include Concept Bottleneck Models \citep{koh2020concept} and Concept Whitening \citep{chen2020concept}. Although most focus has been on interpretability, recent work demonstrates that concept-based auxiliary supervision can significantly improve learning performance. For instance, \citet{fontana2024multitask} show how auxiliary concept tasks enhance computer vision performance through multi-task learning, while \citet{yao2023auxiliary} demonstrate that semantic auxiliary tasks improve retrieval performance even with limited annotations. Similarly, \citet{mahapatra2020structure} integrate semantic guidance into GAN learning, showing improved classification and segmentation performance. These works highlight that concept supervision provides not just interpretability but also serves as a powerful inductive bias that improves sample efficiency and generalization.

Human explanations have proven effective for accelerating learning across multiple reinforcement learning domains. \citet{guan2021widening} demonstrated that factual explanations highlighting important image regions in Human-in-the-Loop RL significantly improve learning speed. Similarly, factual explanations of temporal information in preference-based RL accelerate the learning process \citep{10610505, karlaus2025tempo}. \citet{karalus2022accelerating} showed that integrating counterfactual explanations into the TAMER \citep{knox2009interactively} framework reduces sample complexity. These findings consistently demonstrate that enabling humans to explain different aspects of a task improves both learning speed and sample efficiency.

Our ConceptACT approach is distinguished by its integration of semantic concept learning directly into the action chunking transformer architecture for imitation learning. Unlike methods that discover concepts unsupervised or use them solely for interpretability, we leverage human-provided concept annotations as auxiliary supervision to improve sample efficiency. Furthermore, while most concept-based approaches in RL focus on state abstraction or skill discovery, ConceptACT operates at the episode level, providing semantic guidance that helps the policy learn the underlying task structure more effectively. This positions our work at the intersection of interpretable machine learning and practical robotic imitation learning, addressing both the need for sample-efficient learning and human-understandable decision-making in robotic systems.

\section{Discussion}\label{sec6}

Our experimental results demonstrate that ConceptACT achieves significant improvements in both sample efficiency and learning efficiency compared to standard ACT. The method requires fewer demonstration episodes to reach target performance while accelerating convergence during training, addressing two primary cost factors in practical imitation learning deployments: human demonstration time and computational training resources.

% Only two tasks
While our evaluation focuses on two robotic manipulation tasks, the approach generalizes to any imitation learning domain where meaningful high-level concepts can characterize episodes. Robotic manipulation provides a particularly suitable testbed because object properties, spatial relationships, and task constraints are naturally observable and semantically meaningful to human demonstrators. However, the core principle, leveraging episode-level semantic annotations as auxiliary supervision, extends to domains such as autonomous driving (weather conditions, traffic density), game playing (opponent strategies, map features), or household robotics (room types, object categories).

% Annotation cost
The practical annotation burden of our approach remains minimal due to the structured nature of typical imitation learning data collection. Demonstrators often naturally organize sessions around specific object types, environmental conditions, or task variants, collecting "10 episodes with red cubes," "5 episodes in cluttered environments," or "15 episodes with deformable objects." Our episode-level concept annotations align with this existing structure, requiring minimal additional cognitive load since demonstrators already know the properties of objects they are manipulating and the constraints they are satisfying.

%Timestep level
Our choice of episode-level rather than timestep-level concept annotation represents a deliberate trade-off between annotation cost and temporal specificity. While technically feasible to learn from concepts defined for individual timesteps or trajectory segments, such fine-grained annotation would dramatically increase the labeling burden and potentially introduce inconsistencies. Episode-level concepts capture the semantic invariants that remain constant throughout a demonstration while keeping annotation practical and scalable. For tasks requiring temporal concept variation within episodes, multiple shorter episodes would be more appropriate than fine-grained timestep labeling.

% Concept Transformer Structure is important. Minimal Last
The better performance of Concept Transformer integration compared to simple prediction heads demonstrates that our improvements result not only from providing additional concept information, but specifically from the proposed architectural integration. If concept information alone drove the gains, the prediction head ablation would show similar performance to our full method, yet ConceptACT-Heads does not perform better than standard ACT, even in some cases worse. This stark difference validates that our architectural modification, though minimal since its only one layer, is crucial for effective concept utilization and that effective concept integration requires fundamental changes to how information flows through attention mechanisms rather than simply adding auxiliary prediction losses. Our comparison with LAV-ACT further reinforces this conclusion: LAV-ACT has access to comparable semantic information through language descriptions and employs a significantly larger model architecture, yet ConceptACT achieves better performance with fewer parameters. This suggests that structured integration of semantic information through attention mechanisms provides more effective inductive biases than simply conditioning on unstructured semantic inputs, even when those inputs contain similar conceptual content.

\section{Conclusion}\label{sec7}

We introduced ConceptACT, an extension of Action Chunking with Transformers that incorporates episode-level concepts into imitation learning. By enabling human demonstrators to provide concept annotations during data collection, without requiring this information at deployment, we achieve significant improvements in both sample efficiency and learning speed compared to standard ACT.

Our key finding is that how concepts are integrated matters more than simply having access to them. The Concept Transformer approach, which modifies attention mechanisms to align with human concepts, substantially outperforms both naive concept prediction and language-conditioned models despite using fewer parameters. This demonstrates that structured integration of semantic information through attention provides more effective inductive biases than treating concepts as auxiliary prediction targets or unstructured conditioning inputs.

These results have immediate practical implications: ConceptACT can reduce demonstration requirements by annotating episodes with readily observable properties during data collection. Unlike language-conditioned approaches that impose ongoing deployment costs, our training-only integration maintains standard inference requirements while capturing the benefits of semantic knowledge. This addresses a fundamental bottleneck in deploying imitation learning systems where demonstration data is expensive but semantic expertise is abundant.

While our evaluation focused on manipulation tasks with discrete concepts, the principle of leveraging human semantic understanding during training extends to any domain where episodes exhibit consistent high-level properties. Future work should investigate concept discovery mechanisms to reduce annotation burden and examine how concept-guided learning changes the learned representations to be more interpretable. 

The gap between how humans conceptualize tasks and how robots learn them remains a fundamental challenge in robotics. ConceptACT demonstrates that this gap can be bridged by allowing humans to teach not just through actions, but through the semantic understanding they naturally possess, making robot learning more aligned with humans while remaining practically deployable.

\appendix

\section{Architectural Ablation: Prediction Head Approach}\label{appendix:prediction_heads}

To demonstrate that the architectural integration of concepts through attention mechanisms is crucial for performance gains, rather than simply providing additional concept information, we implement a simpler prediction head baseline. This ablation study treats concept learning as an auxiliary classification task without modifying the core attention mechanism.

\subsection{Architecture Design}

In the evaluation of an architectural variant, the prediction head was used as a baseline, which represents a simple integration of concept information (via an auxiliary loss) without the burden of all the changes of our modified Concept Transformer. To achieve this, we defined the simplest integration which predicts the used concepts (with the help of some MLP layers) as a normal output. This is done on the final encoder representation. Specifically, we use the first token of the encoder output sequence, which serves as a global representation of the current state. This token, denoted as $h_{[CLS]} \in \mathbb{R}^d$, aggregates information from all input modalities (visual features, proprioceptive state, and latent variables) through the self-attention mechanism of the transformer encoder.

For each concept class $T_j$, we define a dedicated prediction head consisting of a multi-layer perceptron:

\begin{equation}
\text{Head}_j = \text{Linear}_{|T_j|}(\text{ReLU}(\text{Linear}_{\text{h}}(h_{[CLS]})))
\end{equation}

where $\text{Linear}_{\text{h}}: \mathbb{R}^d \rightarrow \mathbb{R}^{d_{\text{concept}}}$ projects the encoder representation to a concept-specific hidden dimension $d_{\text{concept}}$, and $\text{Linear}_{|T_j|}: \mathbb{R}^{d_{\text{concept}}} \rightarrow \mathbb{R}^{|T_j|}$ produces logits for the $|T_j|$ concept values within class $T_j$. We include \textit{ReLU} activation and dropout (with probability 0.2) for regularization.

The predicted logits for concept class $T_j$ are thus:

\begin{equation}
\hat{c}^{T_j} = \text{Head}_j(h_{[CLS]}) \in \mathbb{R}^{|T_j|}
\end{equation}

\subsection{Loss Function}

Each concept class is trained independently using cross-entropy loss with label smoothing. For concept class $T_j$ with ground-truth one-hot vector $c^{T_j} \in \{0,1\}^{|T_j|}$, we first convert to class indices:

\begin{equation}
y^{T_j} = \arg\max_k c_k^{T_j}
\end{equation}

The loss for concept class $T_j$ becomes:

\begin{equation}
\mathcal{L}_j^{\text{CE}} = \mathbb{E}\left[-\sum_{k=1}^{|T_j|} q_k^{T_j} \log \text{softmax}(\hat{c}_k^{T_j})\right]
\end{equation}

where $q_k^{T_j}$ represents the label-smoothed target distribution with parameter $\epsilon = 0.1$. The complete concept loss aggregates across all concept classes:

\begin{equation}
\mathcal{L}_{\text{concept}}^{\text{heads}} = \sum_{j=1}^K \mathcal{L}_j^{\text{CE}}
\end{equation}

The total training objective becomes:

\begin{equation}
\mathcal{L}_{\text{total}} = \mathcal{L}_{ACT} + \lambda_{\text{concept}} \mathcal{L}_{\text{concept}}^{\text{heads}}
\end{equation}

where $\mathcal{L}_{ACT}$ includes the standard action reconstruction and KL divergence terms.

\subsection{Ablation Results}

This prediction head approach provides the same concept supervision signal as our primary method but without architectural integration through attention mechanisms. As demonstrated in our experimental results, this approach yields minimal improvement over standard ACT, confirming that effective concept integration requires fundamental changes to information flow through the network rather than simply adding auxiliary prediction losses. The substantial performance difference between this baseline and our Concept Transformer approach validates the importance of architectural design in concept-based learning.

\section{Hyperparameters}\label{appendix:hyperparams}

Tables~\ref{tab:act_hyperparams}, \ref{tab:conceptact_hyperparams}, and \ref{tab:lavact_hyperparams} provide comprehensive hyperparameter specifications for all experimental methods. ACT parameters follow the original implementation from \citet{zhao2023act}.

\begin{table}[ht]
\centering
\caption{Hyperparameters for standard ACT baseline}\label{tab:act_hyperparams}
\begin{tabular}{@{}lc@{}}
\toprule
Parameter & Value \\
\midrule
\textbf{Architecture} & \\
Vision backbone & ResNet-18 \\
Pretrained weights & ImageNet-1K V1 \\
Transformer dimension & 512 \\
Number of attention heads & 16 \\
Feedforward dimension & 3200 \\
Feedforward activation & ReLU \\
Encoder layers & 4 \\
Decoder layers & 1 \\
Dropout rate & 0.1 \\
\textbf{Action Chunking} & \\
Chunk size & 100 \\
Action steps per invocation & 100 \\
Observation steps & 1 \\
\textbf{VAE Configuration} & \\
Use VAE & True \\
Latent dimension & 32 \\
VAE encoder layers & 4 \\
KL divergence weight & 10.0 \\
\textbf{Training} & \\
Learning rate &  $3 \times 10^{-5}$ \\
Batch size & 8 \\
Training epochs & 10 \\
Maximum steps & 30,000 \\
Weight decay & $1 \times 10^{-4}$\\
Backbone learning rate & $1 \times 10^{-5}$ \\
\bottomrule
\end{tabular}
\end{table}

\begin{table}[ht]
\centering
\caption{Hyperparameters for ConceptACT with class-aware concept transformer}\label{tab:conceptact_hyperparams}
\begin{tabular}{@{}lc@{}}
\toprule
Parameter & Value \\
\midrule
\textbf{Concept-Specific Parameters} & \\
Concept learning enabled & True \\
Concept integration method & Transformer (cross-entropy) \\
Concept dimension & 128 \\
Concept loss weight ($\lambda$) & 0.2 \\
\textbf{Concept Types for Task 1} & \\
Object shape classes & 3 \\
Object color classes & 4 \\
Target class & 2 \\
\textbf{Concept Types for Task 2} & \\
Object 0 shape classes & 3 \\
Object 0 color classes & 4 \\
Object 1 shape classes & 3 \\
Object 1 color classes & 4 \\
Object 2 shape classes & 3 \\
Object 2 color classes & 4 \\
Ordering classes & 3 \\
\textbf{Training} & \\
Learning rate &  $3 \times 10^{-5}$ \\
Batch size & 8 \\
Training epochs & 10 \\
\midrule
\multicolumn{2}{c}{Other parameters inherited from ACT} \\
\bottomrule
\end{tabular}
\end{table}

\begin{table}[ht]
\centering
\caption{Hyperparameters for LAV-ACT with language conditioning}\label{tab:lavact_hyperparams}
\begin{tabular}{@{}lc@{}}
\toprule
Parameter & Value \\
\midrule
\textbf{Language-Specific Parameters} & \\
Voltron model variant & v-cond \\
FiLM hidden dimension & 512 \\
Voltron encoder frozen & True \\
\textbf{Training} & \\
Learning rate &  $3 \times 10^{-5}$ \\
Batch size & 8 \\
Training epochs & 10 \\
Maximum steps & 30,000 \\
Weight decay & $1 \times 10^{-4}$\\
Backbone learning rate & $1 \times 10^{-5}$ \\
\midrule
\multicolumn{2}{c}{Other parameters inherited from ACT} \\
\bottomrule
\end{tabular}
\end{table}

\subsection{Evaluation Configuration}

All methods were evaluated using identical protocols: 10 test episodes per evaluation checkpoint, assessment every full epoch, and experiments conducted across 10 random seeds (42, 123, 456, 100, 101, 102, 103, 104, 105, 106).

\subsection{Number of Parameters}\label{appendix:parameters}

See table \ref{tab:parameter_numbers} to compare the amount of parameters each methods has in the respective task. 

\begin{table}[ht]
    \centering
    \begin{tabular}{l|r|r}
       Method  &  Task 1 & Task 2 \\
       \midrule
       ACT  & 51.6M  & 51.6M\\
       ConceptACT - Transformer  & 56.1M & 61.3M\\
       ConceptACT - Heads  & 51.8M & 52.1M\\
       LAV-ACT  & 151.8M & 151.8M\\

    \end{tabular}
    \caption{Number of parameters for each model type in each task.}
    \label{tab:parameter_numbers}
\end{table}

\section{Detailed Task descriptions}

\subsection{Task 1 (Sorting):}\label{appendix:task1}
Pick-and-place operations with conditional sorting constraints. The robot must grasp objects of varying shapes and colors from randomized positions and sort them into two designated collection areas based on logical rules combining object properties.

\textbf{Object Space:} The available objects are defined by shape-specific color constraints:
\begin{align}
S &= \{\text{cube}, \text{rectangle}, \text{cylinder}\} \\
C_{\text{cube}} &= \{\text{red}, \text{green}, \text{yellow}\} \\
C_{\text{rectangle}} &= \{\text{red}, \text{blue}, \text{green}, \text{yellow}\} \\
C_{\text{cylinder}} &= \{\text{red}, \text{blue}, \text{green}\} \\
L &= \{1, 2, 3, 4, 5\} \text{ (pickup locations)}
\end{align}

\textbf{Sorting Rule:} Given an object with color $c$ and shape $s$, the target collection area is determined by:
\begin{equation}
\text{Target} = \begin{cases}
\text{Area A} & \text{if } (\text{shape} = \text{cube} \\ & \land \text{color} \in \{\text{red}, \text{green}\}) \\
& \text{or } \\
& (\text{shape} = \text{cylinder} \\
& \land \text{color} = \text{blue}) \\
\text{Area B} & \text{otherwise}
\end{cases}
\end{equation}

The idea of this rule is to create a complex logical disjunction where objects are only sorted into Area A if they have two (or more) desired attributes: (1) red coloration regardless of shape, or (2) rectangular shape with any non-yellow color. All remaining objects (yellow rectangles, non-red cubes, and non-red cylinders) are sorted to Area B.

\textbf{Experimental Design:} 
In total, there are 42 valid combinations of possible object shape, color, and location. For the evaluation, 10 combinations were saved (red cubes and yellow rectangles across all 5 locations). The remaining 32 combinations were used for training. This 80/20 split ensures evaluation includes objects that test both branches of the sorting rule.

\textbf{Evaluation Metric:} Hierarchical scoring system $\in [0, 3]$ where: 0 (failed grasp), 1 (successful grasp, failed placement), 2 (correct placement, incorrect sorting area), 3 (complete success with correct sorting according to the rule).

\subsection{Task 2 (Ordering):}
\label{appendix:task2}
The second task is about sequential placement of three different objects, at three different pick-up zones into a single collection area. Crucial in this task is the order of the placement into the area, which is dependent on the available objects and their properties.

\textbf{Object Space:} The available objects are defined by shape-color combinations:
\begin{align}
S &= \{\text{cube}, \text{rectangle}, \text{cylinder}\} \\
C_{\text{cube}} &= \{\text{red}, \text{green}, \text{yellow}\} \\
C_{\text{rectangle}} &= \{\text{red}, \text{blue}, \text{green}, \text{yellow}\} \\
C_{\text{cylinder}} &= \{\text{red}, \text{blue}, \text{green}\}
\end{align}

\textbf{Constraint System:} Let $\mathbf{o} = [o_1, o_2, o_3]$ be a bottom-to-top ordering sequence. Sequences must fulfill the following constraints to be valid:

\begin{enumerate}

\item \textbf{Cardinality constraints:} 
\begin{itemize}
    \item $|\{\text{rectangle}\}| \geq 1$
    \item $|\{\text{rectangle}\}| \leq 2$
    \item $|\{\text{cube}\}| \leq 2$
    \item $|\{\text{cylinder}\}| \leq 1$
\end{itemize}

\item \textbf{Positional constraints:} 
\begin{align}
\text{cylinder} &\Rightarrow \text{position} = 3  \\
\text{rectangle} &\Rightarrow \text{position} \in \{1, 2\}
\end{align}

\item \textbf{Color hierarchy:} \\
$\{\text{red}, \text{green}\} \prec \text{yellow} \prec \text{blue}$

\item \textbf{Adjacency rule:} Objects with identical colors must be vertically adjacent when multiple instances exist
\end{enumerate}

These constrains reduced the theoretical full combinatorial spaces from $10^3 = 1000$ possible sequences to exactly 26 valid configurations. Out of these, 20 configurations are used for the training, while the 6 held-out configurations are used for the evaluation. 

\textbf{Evaluation Metric:} Success score $\in [0, 6]$ where each correctly placed object in the proper sequence position contributes 2 points, with deductions for constraint violations or placement failures.

\bibliographystyle{IEEEtranN}
\bibliography{main}

\end{document}